%% file: Cai_WCL2026-0541.tex
\documentclass[10pt,conference]{IEEEtran}
\IEEEoverridecommandlockouts
\usepackage{cite}
\usepackage{amsmath,amssymb,amsfonts}
\usepackage{algorithm}
\usepackage{graphicx}
\usepackage{epstopdf}
\usepackage{subfig}
\usepackage{epsfig}
\usepackage{textcomp}
\usepackage{xcolor}
\usepackage{algorithm}
\usepackage{algpseudocode}
\usepackage{epstopdf}
\usepackage{subfloat} 
\usepackage{caption}
\usepackage{stfloats}
\usepackage{hyperref}
\usepackage{amsmath}
\usepackage{float}
\usepackage{bm}
\usepackage{gensymb}
\usepackage{multirow}
\usepackage{makecell}
\usepackage{siunitx}
\usepackage[letterpaper, top=0.701in, bottom=1.1in, left=0.625in, right=0.625in]{geometry}
\usepackage{acro}

\input{acronyms.tex}
\def\BibTeX{{\rm B\kern-.05em{\sc i\kern-.025em b}\kern-.08em
    T\kern-.1667em\lower.7ex\hbox{E}\kern-.125emX}}

\begin{document}

\title{{SPOT: Single-Shot Positioning via Trainable Near-Field Rainbow Beamforming}
\thanks{This work was supported by the National Natural Science Foundation of China (NSFC) under Grant 62125108 and by the Science and Technology Commission Foundation of Shanghai under Grant 25DP1500100. The source code is available at \url{https://github.com/SJTU-WirelessAI-Lab/SPOT}.}
}

\author{\IEEEauthorblockN{Yeyue Cai, Jianhua Mo,~\IEEEmembership{Senior Member,~IEEE}, and Meixia Tao,~\IEEEmembership{Fellow,~IEEE}}
\thanks{The authors are with the School of Information Science and Electronic Engineering at Shanghai Jiao Tong University, Shanghai 200240, China (Email:\{caiyeyue, mjh, mxtao\}@sjtu.edu.cn). Corresponding authors: \textit{Meixia Tao} and \textit{Jianhua Mo.}}
}
\maketitle

\begin{abstract}
Phase–time arrays, which integrate phase shifters (PSs) and true‑time delays (TTDs), have emerged as a cost-effective architecture for generating frequency-dependent rainbow beams in wideband sensing and localization. This paper proposes an end-to-end deep learning-based scheme that simultaneously designs the rainbow beams and estimates user positions. Treating PS and TTD coefficients as trainable variables allows the network to synthesize task‑oriented beams that maximize localization accuracy. A lightweight fully connected module then recovers the user’s angle‑range coordinates from its feedback of the maximum quantized received power and its corresponding subcarrier index after a single downlink transmission. Compared with existing analytical and learning‑based schemes, the proposed method reduces overhead by an order of magnitude and delivers consistently lower two‑dimensional positioning error.

\end{abstract}
\begin{IEEEkeywords}
phase-time array, rainbow beam, 
localization.
\end{IEEEkeywords}

\section{Introduction}

\ac{ISAC} combines radar and communication by sharing hardware and spectral resources, making it a key technology for 5G-Advanced and 6G networks \cite{GonzalezPrelcic2025six}. 
A key advantage of \ac{ISAC} is its ability to repurpose communication waveforms, like \ac{OFDM}, for user positioning, enabling dynamic beam steering in subsequent transmissions.

In wideband systems, \acp{PTA}, integrating \acp{PS} and \acp{TTD}, have become a promising architecture for cost-effective and energy-efficient localization \cite{ratnam2022joint,cai2025frequency,Mo2024beamforming, cai2025hybrid}. \acp{PTA} synthesize a controllable ``rainbow beam" by simultaneously directing different \ac{OFDM} subcarriers towards distinct spatial directions. This unique feature facilitates rapid scanning and training across broad angular sectors and distance ranges. In contrast, conventional phased arrays generate only a single directional beam per time slot, necessitating sequential beam sweeping spanning multiple time slots. Consequently, \acp{PTA} significantly reduce beamforming latency and system overhead in wideband \ac{ISAC} applications, attracting considerable research interest.

One of the primary applications of \acp{PTA} in \ac{ISAC} systems is user positioning \cite{gao2023integrated,luo2024beam,lei2025deep,zheng2025near}. Within this framework, the \ac{BS} transmits rainbow-like pilot beams covering the entire service area and subsequently estimates user locations based on feedback from the users. In far-field scenarios, rainbow-beam training schemes have been proposed to recover user angles with reduced overhead \cite{gao2023integrated}. Here, the \ac{BS} estimates each user’s direction by mapping the reported subcarrier frequency to its corresponding steering angle. In near-field environments, where spherical wavefronts become significant, rainbow beams are generated separately in angle and distance to enable sequential estimation of both parameters through a controllable beam squint (CBS) scheme \cite{luo2024beam}. Convolutional neural networks are then employed to refine these initial estimates using the received signal \cite{lei2025deep}. A distance-dependent rainbow-beam method further reduces overhead by enabling joint angle and distance estimation within a beam \cite{zheng2025near}. In contrast to the aforementioned cooperative approaches, a deep learning-based uplink near-field positioning method, called RaiNet, was proposed in \cite{Klus2025Deep}. The \ac{BS} observes the full-band amplitude response vector, and a convolutional backbone regresses the 2D position in a single shot.

As discussed above, existing PTA-based sensing methods in wideband systems predominantly rely on rainbow beams that map frequency to either angle or range \cite{gao2023integrated,luo2024beam,lei2025deep}. However, range estimation via frequency-distance mapping is effective primarily in the near-field where the beam can be focused on a specific point. Moreover, these schemes often suffer from high feedback overhead. For instance, studies \cite{gao2023integrated} and \cite{luo2024beam} typically require at least two rounds of downlink beam transmissions and subsequent user feedback to recover both angle and range. The approach presented in \cite{lei2025deep} further exacerbates this overhead by requiring users to transmit their complete received signal in each feedback instance. The method in \cite{zheng2025near} can scan the range and angle simultaneously, but multiple beams are needed to cover the whole angle-range space and avoid coverage gaps. Moreover, this rainbow-beam design typically covers a wide sector of \(-90^{\circ}\) to \(90^{\circ}\), which is broader than the actual user distribution range (e.g., \(-60^{\circ}\) to \(60^{\circ}\)), leading to unnecessary beam coverage and resource waste.


To overcome these limitations, we propose a deep learning-based localization scheme capable of recovering both angle and range with only a single downlink beamforming transmission and one-shot user feedback message, referred to as Single-shot POsitioning with Trainable near-field rainbow beamforming (SPOT). In the SPOT scheme, each user identifies the \ac{OFDM} subcarrier with the maximum received power and reports solely the quantized maximum received power and the index of the corresponding subcarrier. The \ac{BS} then infers the user’s range and angle based on the user feedback. Furthermore, to ensure robust performance over a wide range of user distances, we deviate from prior works that determine \ac{PS} and \ac{TTD} values heuristically. Instead, we treat the \ac{PS} and \ac{TTD} coefficients as learnable network parameters. This joint optimization process yields rainbow beams that are inherently optimized for accurate position estimation across all propagation regimes.

\section{System Model}
\begin{figure}[t]    \centering    \includegraphics[width=0.7\linewidth]{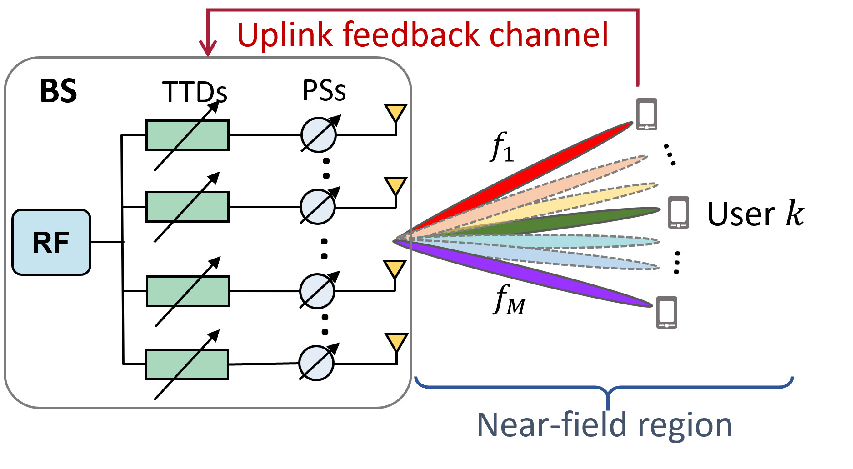}
    \caption{System model.}
    \label{fig: PTA}
    \vspace{-1.5em}
\end{figure}

We consider an \ac{OFDM} wideband \ac{ISAC} system, where the \ac{BS} is equipped with a single \ac{RF} chain connected to a uniform linear array (ULA) consisting of $N$ elements. As shown in Fig. \ref{fig: PTA}, we employ a \ac{PTA} architecture to generate rainbow-like beams for radar sensing of \(K\) users. The \ac{LoS} channel model is assumed for all $K$ users.
In the \ac{PTA} architecture, the single \ac{RF} chain is connected to all \(N\) transmit antennas via \(N\) \acp{PS} and \(N\) \acp{TTD}. The system utilizes $M$ orthogonal subcarriers. Let $B$ denote the system bandwidth and $f_c$ the central carrier frequency. Accordingly, subcarrier \(m\) has a center frequency of $f_m=f_c+\frac{(2m-1-M)}{2}f_{\mathrm{scs}}, \forall m \in \{1, 2, \dots, M\} $, where subcarrier spacing \(f_{\mathrm{scs}}= \frac{B}{M}\). 

The position of user \(k\) relative to the center of the ULA at the \ac{BS} is characterized by the angle $\phi_{k}$ and distance $r_{k}$. The coordinates of user \(k\) and the $n$-th element of the ULA are denoted, respectively, by $\mathbf{u}_{k}=\left[r_{k} \cos \phi_{k}, r_{k} \sin \phi_{k}\right]^{T}$ and $\mathbf{c}_{n}=\left[x_{n} d, 0\right]^{\mathrm{T}}, \forall n \in \{1, 2, \dots, N\}$, where $x_{n}=n-\frac{N+1}{2}$ and antenna spacing $
d=\frac{c}{2f_c}$. 
In the near-field region, the propagation distance from the $n$-th antenna element to user $k$ is approximated following the spherical wave model as 
\begin{equation}
r_{k, n}=\left\|\boldsymbol{u}_{k}-\boldsymbol{c}_{n}\right\|{\approx} r_{k}-x_{n} d \cos \phi_k +\frac{x_{n}^2 d^2 \sin^2{\phi_k}}{2r_k}.
\end{equation}

The \ac{LoS} channel for subcarrier \(m\) of user $k$ is modeled as 
\begin{equation}
\mathbf{h}_{k,m} = \beta_{k,m}e^{j\frac{2\pi f_m r_k}{c}} \mathbf{a}_{k,m},
\end{equation}
where $\beta_{k,m}=\frac{c}{4\pi f_mr_k}$ denotes the path loss, and $\mathbf{a}_{k,m}$ is the array response vector, which is given by
\begin{equation}
\mathbf{a}_{k,m} \!=\! \left[e^{-j \frac{2 \pi f_m}{c}\left(r_{k, 1}-r_{k}\right)}, \ldots, e^{-j \frac{2 \pi f_m}{c}\left(r_{k, N}-r_{k}\right)}\right]^{T}.
\end{equation}

The BS transmits a sensing signal \(s_m\) on each subcarrier using the \ac{PTA}-induced wideband beamforming weights. The received signal of user \(k\) on subcarrier \(m\) is
\begin{equation}
\begin{split}
\label{eq: receive}
y_{k,m} &= \mathbf{h}_{k,m}^H  e^{j\left(\mathbf{\Phi} - 2 \pi f_{m} \mathbf{T}\right)} s_{m}  + n_{k,m}, 
\end{split}
\end{equation}
where \(n_{k,m}\) denotes the additive noise. Vectors $\mathbf{\Phi}\in \mathbb{R}^{N \times 1}$ and $\mathbf{T}\in \mathbb{R}^{N \times 1}$ correspond to the phase shifts and time delays implemented by \acp{PS} and \acp{TTD}, respectively. After this single downlink shot, each user computes the received power spectrum \(\{|y_{k,m}|^2\}_{m=1}^{M}\) and feeds back a compact message consisting of the peak power and the corresponding subcarrier index. Upon collecting the feedback from all users, the BS estimates the angular and range parameters \(\{\phi_k,r_k\}_{k=1}^{K}\). Our goal is to optimize the PTA beamforming characterized by $\mathbf{\Phi}$ and $\mathbf{T}$, together with the BS-side estimation function, to recover \(
\phi_k\) and \(r_k\) for all the $K$ users.


\section{Single-Shot Positioning via Trainable Near Field Rainbow Beamforming}
\begin{figure}[t]    
\centering    \includegraphics[width=0.95\linewidth]{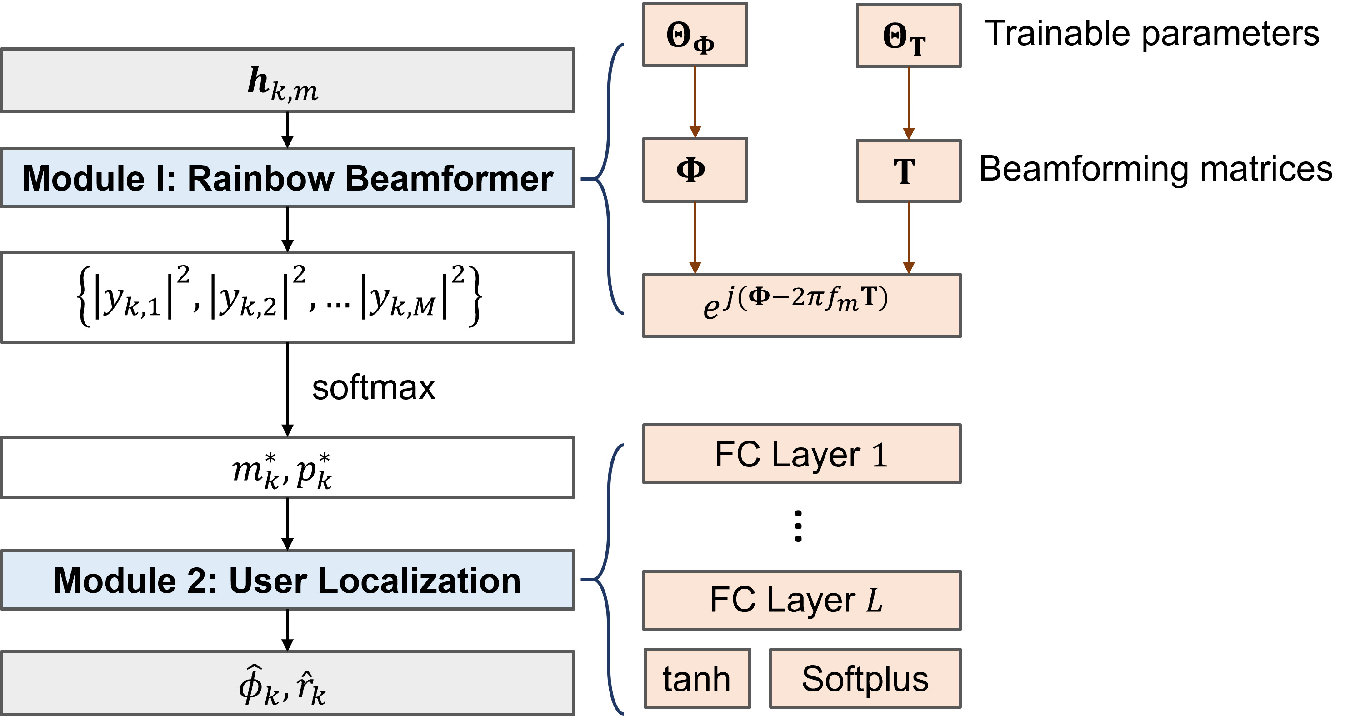}
    \caption{System architecture of the proposed SPOT scheme.}
    \label{fig: NN}
    \vspace{-1.5em}
\end{figure}

This section presents the proposed SPOT scheme for single-shot near-field localization. As shown in Fig. \ref{fig: NN}, SPOT consists of two jointly trained modules: a trainable beamformer that generates the downlink wideband rainbow beam via learnable \ac{PS} and \ac{TTD} settings and an estimator deep neural network (DNN) that maps the uplink feedback to the user angle and range. The two modules are optimized end-to-end to minimize the localization RMSE.


\subsection{Module 1: Rainbow Beam Design}
Module 1 parameterizes the PTA by learnable $\boldsymbol{\Theta}_{\Phi}\in\mathbb{R}^{N\times 1}$ and $\boldsymbol{\Theta}_{T}\in\mathbb{R}^{N\times 1}$. Given channel realizations $\{\mathbf{h}_{k,m}\}_{k=1,\ldots,K;\,m=1,\ldots,M}$ during training, it learns task-oriented $\mathbf{\Phi}$ and $\mathbf{T}$ for single-shot positioning.

Upon receiving the downlink signal $y_{k,m}$, the user extracts features for localization. To improve numerical stability and compress the dynamic range of received power across different signal-to-noise ratios, we first convert the received signal power into the logarithmic domain, $P_{k,m} = 10\log_{10}(|y_{k,m}|^2)$.

To enable end-to-end gradient backpropagation, we replace the non-differentiable $\arg\max(\cdot)$ operation during training with a temperature-scaled softmax. The weight for each subcarrier is computed as:
\begin{equation}
w_{k,m} = \frac{\exp(\alpha P_{k,m})}{\sum_{m'=1}^{M} \exp(\alpha P_{k,m'})},
\end{equation}
where $\alpha > 0$ is the temperature parameter. Based on numerical sensitivity analysis, we set $\alpha=100$ to ensure the selection is sufficiently discriminative while maintaining smooth gradient flow. The differentiable soft subcarrier index $m_k^{\star}$ and peak-power feature $ p_k^{\star}$ are then obtained via $m_k^{\star} = \sum_{m=1}^{M} w_{k,m} m$ and $ p_k^{\star} = \sum_{m=1}^{M} w_{k,m} P_{k,m}$.

\subsection{Module 2: User Localization}
The position estimation process is implemented using a DNN, which takes the maximum received power \( p_k^{\star}\) and the corresponding subcarrier index \(m_k^{\star}\) as input. The input features are processed through three fully connected (FC) layers, where the output of each layer is denoted as \(\mathbf{v}_i\) for \(1\leq i \leq L\). To constrain the angle $\hat{\phi}_k \in [-\pi/3, \pi/3]$ and ensure distance $\hat{r}_k > 0$, activation functions are applied \footnote{The angular range $\phi_k \in [-\pi/3, \pi/3]$ follows the standard $120^\circ$ sectorization widely adopted in cellular deployments.}:
\begin{equation}
    \hat{\phi}_k = \frac{\pi}{3} \tanh(\mathbf{v}_{L,1}), \quad \hat{r}_k = \text{Softplus}(\mathbf{v}_{L,2}),
\end{equation}
where $\mathbf{v}_L = [\mathbf{v}_{L,1}, \mathbf{v}_{L,2}]$ is the final DNN output.

\subsection{Loss Function and Computational Complexity}
The network is trained using a loss function defined as the RMSE of 2D localization
\begin{equation}
\mathcal{L} = \sqrt{\frac{1}{K} \sum_{k=1}^K \left[ (\hat{x}_k - x_k)^2 + (\hat{y}_k - y_k)^2 \right]},
\end{equation}
where $\hat{x}_k = \hat{r}_k \cos \hat{\phi}_k$ and $\hat{y}_k = \hat{r}_k \sin \hat{\phi}_k$ denote the estimated Cartesian coordinates.

The proposed network's main computational cost lies in Module 2, a DNN with $L=4$ FC layers. With $N_0 = 2$ input neurons (power and subcarrier index), $N_L = 2$ output neurons (angle and distance), and hidden layers $N_1$ to $N_{L-1}$, its per-user complexity is $\mathcal{O}\left( \sum_{\ell=0}^{L-1} N_{\ell} N_{\ell+1} \right)$.

\subsection{Deployment Issues and Discussions}
During deployment, the differentiable soft subcarrier index is replaced by a hard decision for each user. Each user reports \(m_k^{\star}=\arg\max_{m} P_{k,m}\) together with the quantized peak power. Specifically, the peak power is first converted to the logarithmic domain and then uniformly quantized with \(b\) bits over a dynamic range \([P_{\min},P_{\max}]\), where \(P_{\min}\) and \(P_{\max}\) are set offline from training-set peak-power percentiles for the target deployment geometry and reused for all users within the same scenario. In addition, the learned parameters are projected onto their physical ranges via $\boldsymbol{\Phi}={\mathrm{mod}}\bigl(\boldsymbol{\Theta}_{\Phi},\,2\pi\bigr)$ and $\mathbf{T}={\mathrm{mod}}\bigl(\boldsymbol{\Theta}_{T}-\min{(\boldsymbol{\Theta}_{T})},\,1/f_{\mathrm{scs}}\bigr)$. Note that such projection will not affect the performance due to the periodic nature of the array response in both phase and delay.

This work assumes ideal PS and TTD control, whereas practical PS/TTD hardware has finite resolution. In particular, TTD resolution is key to preserving the rainbow-beam frequency-to-angle mapping. Future work will study quantization-aware and robust training with priors or regularization to improve robustness and interpretability under coarse PS/TTD and more diverse deployment scenarios.

\section{Numerical Results}

Unless explicitly stated otherwise, the simulations are conducted using the parameters detailed in Table \ref{para}. It is worth noting that the Rayleigh distance for our simulation setup is \(348.35\) meters. In our simulations, $K$ users are assumed to be uniformly and randomly distributed within a spatial sector spanning \([-60^\circ, 60^\circ]\) relative to the BS, with distances ranging from 5 to 300 meters. For the purpose of model development and evaluation, a dataset comprising \num{50 000} samples is generated for the training set, while \num{5 000} independent samples are generated for the validation and test sets, respectively.

\begin{table}[t]
\centering
\caption{Main notations and their typical values.}
\label{para}
\begin{tabular}{|c| c| c|}
\hline
\textbf{Notation} &\textbf{Definition} & \textbf{Value} \\
\hline
$f_c$ &  { Central carrier frequency } & 28 GHz \\
\hline
$B$ &  {Total bandwidth} & 380.16 MHz \\
\hline
$f_{\text{scs}}$ &  {Subcarrier spacing} & 240 kHz \\
\hline
$N$ &  { Number of antennas at BS} & 256 \\
\hline
$M$ &  { Number of subcarriers} & 1584 \\
\hline
$\sigma^2$ &  {Thermal noise power density} & -174 dBm/Hz \\
\hline
$P_t$ &  { Transmit power at BS} &  40 dBm \\
\hline
$N_0, N_1, N_2, N_3, N_4$ &  {DNN parameters} & 2, 64, 128, 64, 2 \\
\hline
\end{tabular}
\vspace{-1.2em}
\end{table}



For comparison, we consider the following baselines:
\begin{itemize}
  \item \textbf{CBS}~\cite{luo2024beam}: 
  Two stages of downlink transmissions include one beam for angle (CBS Beam 1) and $K$ user-specific beams for distance (CBS Beam 2) estimation, respectively. Each user feeds back its own peak-power subcarrier index, enabling the \ac{BS} to infer the position.

  \item \textbf{ConvNeXt}~\cite{lei2025deep}: This scheme first runs CBS to get a coarse position estimate, then refines it with a ConvNeXt model that takes the two received power vectors and the coarse estimate as input.
  \item \textbf{Distance-dependent CBS}~\cite{zheng2025near}: This scheme uses distance-aware rainbow beams for joint angle-distance estimation within a single beam. Multiple beams are employed to improve the accuracy.
  \item \textbf{RaiNet} \cite{Klus2025Deep}: We adapt the uplink-based RaiNet to our downlink scenario where the \ac{BS} transmits CBS rainbow beams, and each user feeds back its received power vector across subcarriers. The \ac{BS} employs the original RaiNet which includes 3 convolutional layers and 3 FC layers to regress the user's position.
\end{itemize}



\subsection{2D RMSE under Different Maximum User Ranges}
\begin{figure}[t]    
    \centering    
    \includegraphics[width=1.00\linewidth]{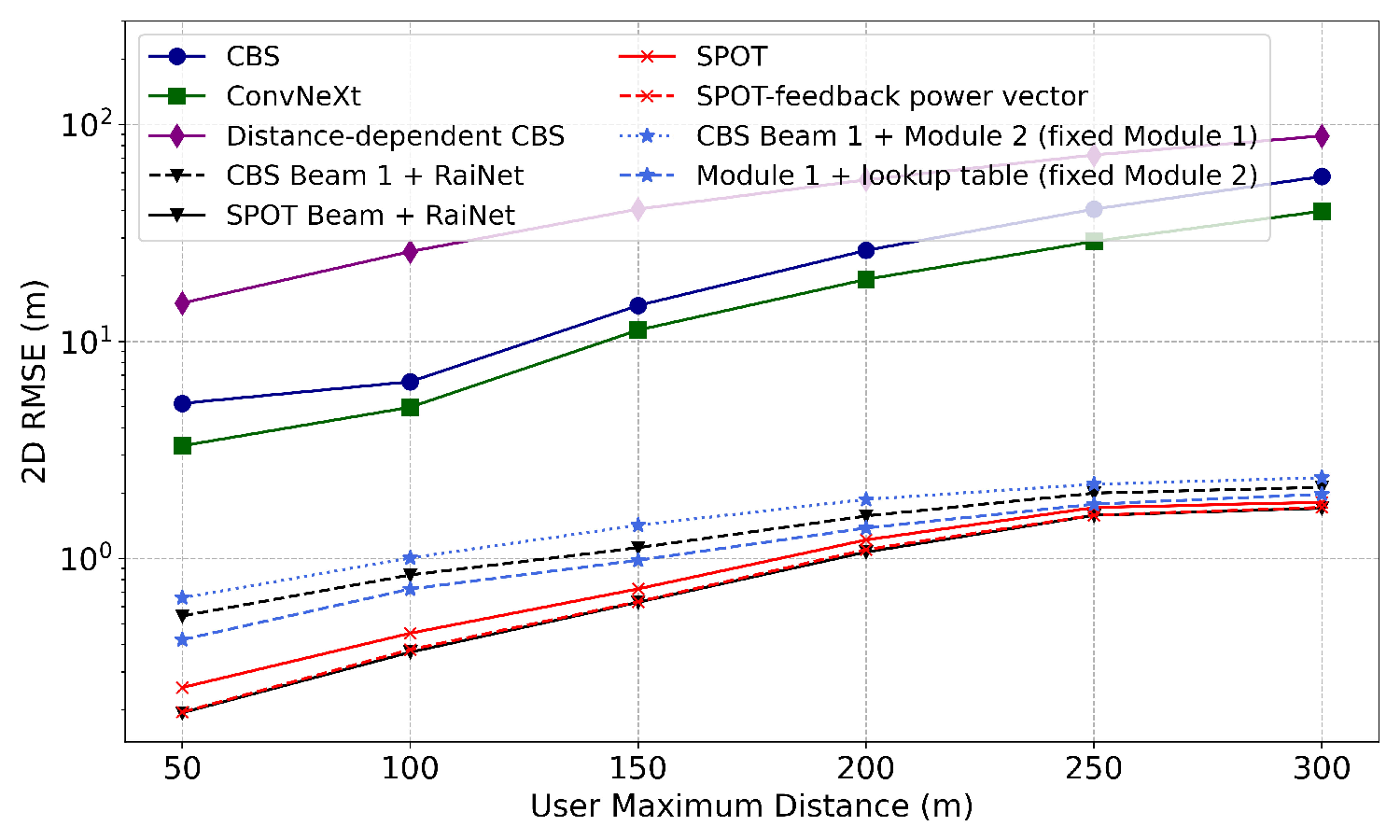}
    \caption{2D RMSE under different user maximum distances.}
    \label{fig: dis_range}
\vspace{-1.5em}
\end{figure}

Fig.~\ref{fig: dis_range} reports the 2D localization RMSE versus the maximum user distance. The CBS, distance-dependent CBS, and ConvNeXt baselines all exhibit consistently higher errors than the SPOT scheme, and their performance deteriorates more rapidly as the operating range expands. This behavior can be traced to two limitations of these baselines. First, as will be shown in Fig. \ref{fig: gain}, the near-field focusing effect on which they depend weakens quickly with distance, so the subcarrier-index-to-distance mapping becomes coarse for far users, leading to a noticeable loss of ranging resolution. Second, since they follow a two-stage procedure in which distance is inferred conditionally on the previously estimated angle, these methods are vulnerable to error propagation, i.e., the first-stage angular error is amplified in the second-stage distance estimate.

Fig. \ref{fig: dis_range} compares several variants of the SPOT and RaiNet schemes, which are both one-stage methods.
First, the proposed SPOT achieves lower RMSE than the original CBS-based RaiNet method in \cite{Klus2025Deep} (`CBS Beam 1 + RaiNet'). 
Second, replacing the RaiNet downlink beam with the trained SPOT beam (`SPOT Beam + RaiNet') reduces the RMSE by approximately 39.5\%. This variant also outperforms the original SPOT scheme that feeds back only the maximum power and its subcarrier index. 
Additionally, when the SPOT input is changed to the full $M$-dimensional received power vector (`SPOT-feedback power vector'), the performance becomes comparable to SPOT beam-based RaiNet while using a much smaller model. The computational cost is reduced by 30 times, requiring only 0.65 M FLOPs compared to 19.37 M for RaiNet.

Finally, an ablation study was conducted to validate the contributions of the two SPOT modules. When the beamforming module (Module 1) is fixed to the CBS beam while only the estimator (Module 2) is trained, the overall localization error is noticeably higher. Allowing the beamformer to be trainable lowers the average 2D RMSE by approximately 40.9\% relative to this fixed baseline, underscoring the central role of beam optimization. Conversely, replacing the learned estimator (Module 2) with a non-learned simple distance lookup method increases the 2D RMSE by 31.2\%. 
This lookup table is constructed by sampling power levels at various distances. 
This result validates the critical importance of the learned estimator for refining distance inference.

\subsection{Rainbow Beam Patterns}
\begin{figure}[t]
    \centering
    \subfloat[]{
    \includegraphics[width=0.32\linewidth]{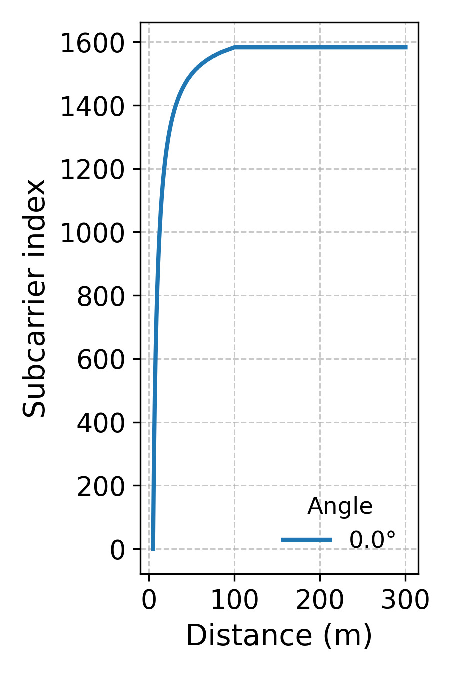}
    \label{fig: distance_index_mapping}
    }
    \hfil
    \subfloat[]{
    \includegraphics[width=0.49\linewidth]{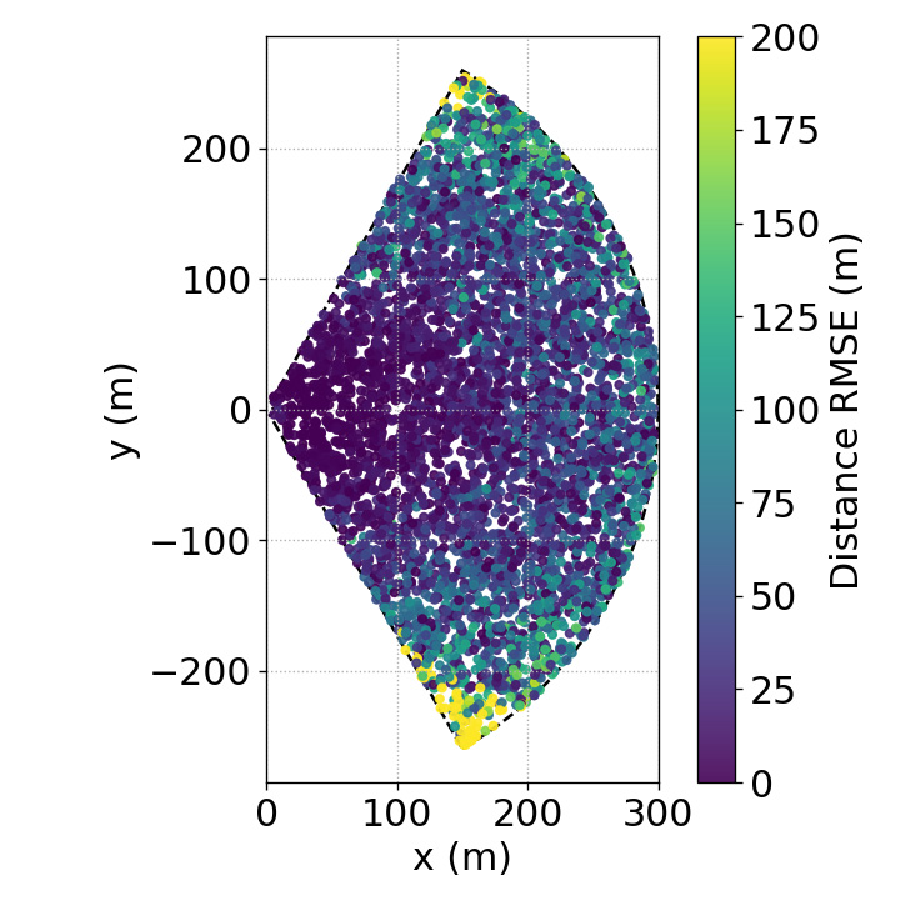}
    \label{fig: distance_rmse}}
    \caption{Distance estimation performance of CBS Beam 2. (a) Subcarrier index vs. distance mapping for a rainbow beam scanning from $(0^{\circ}, 5 \text{ m})$ to $(0^{\circ}, 300 \text{ m})$. (b) Distance RMSE.}
    \label{fig: gain}
\vspace{-1em}
\end{figure}

To explain the performance gain of our method over the CBS baseline, we illustrate in Fig.~\ref{fig: gain} how the subcarrier index of CBS Beam 2 maps to the distance. As the user moves farther away, near-field focusing weakens and the per-subcarrier beams broaden, which compresses the resolvable distance intervals in Fig.~\ref{fig: distance_index_mapping}, and leads to degraded ranging accuracy at long ranges in Fig.~\ref{fig: distance_rmse}.

\begin{figure}[t]
    \centering
    \subfloat[CBS Beam 1 (5--50 m)]{
        \includegraphics[width=0.4\linewidth]{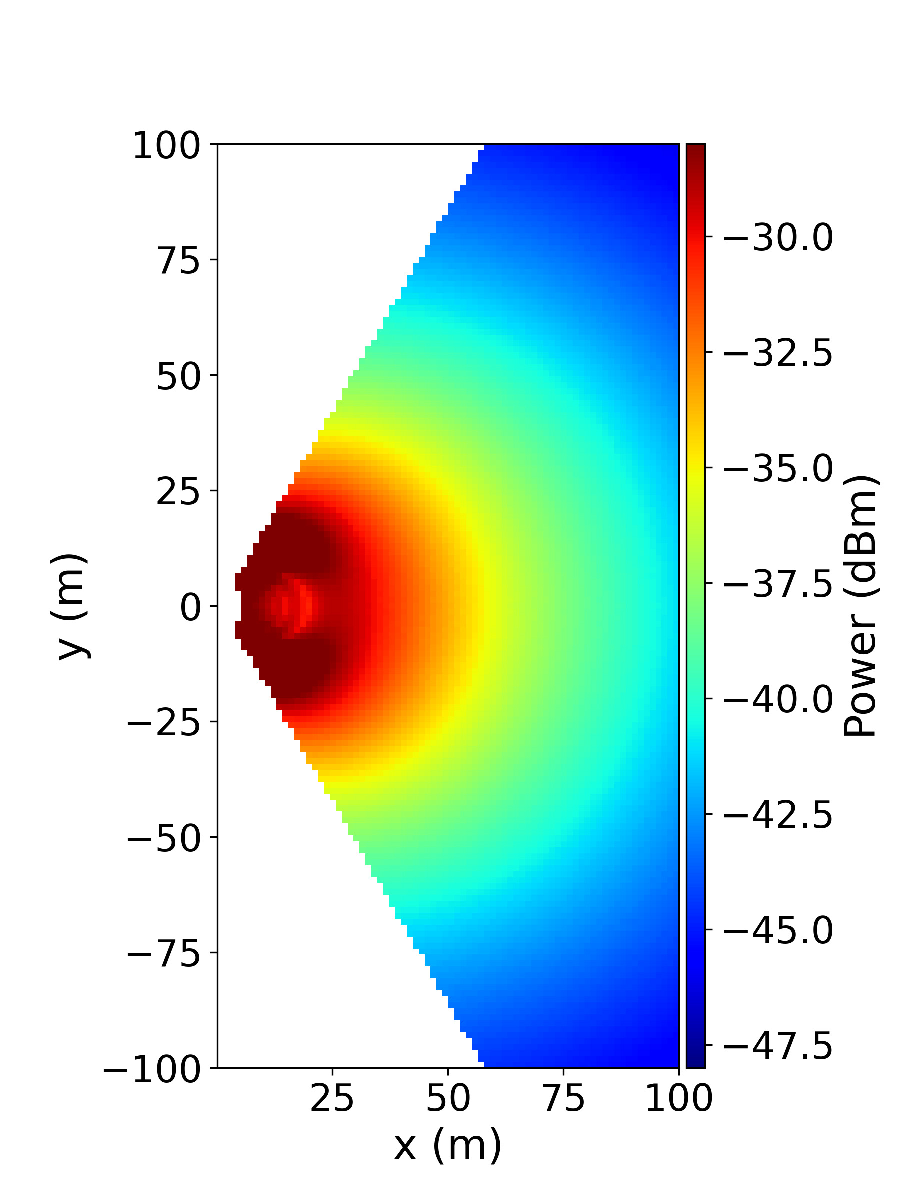}
        \label{fig: cbs_5_50}
    }
    \hfil
    \subfloat[CBS Beam 1 (5--300 m)]{
        \includegraphics[width=0.4\linewidth]{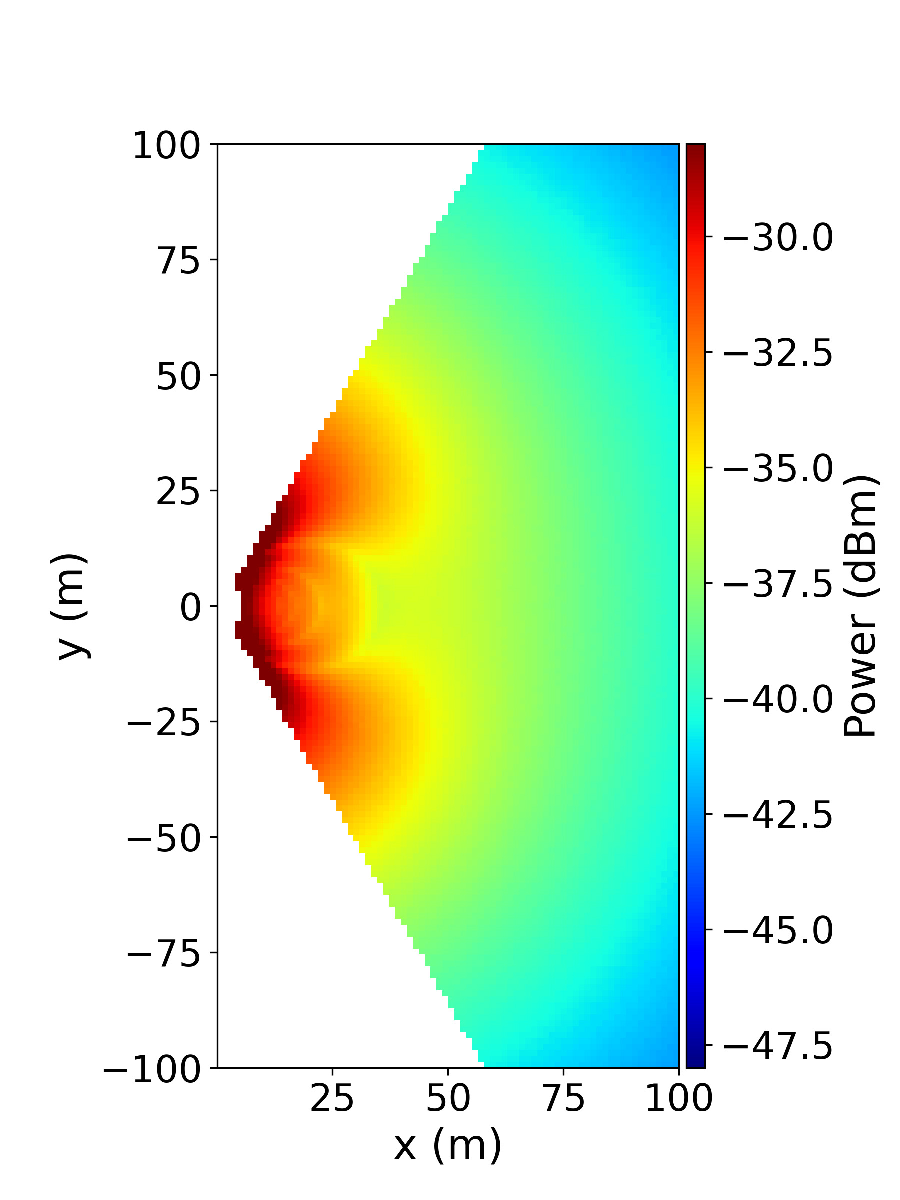}
        \label{fig: cbs_5_300}
    }
    \hfil
    \subfloat[SPOT beam (5--50 m)]{
        \includegraphics[width=0.4\linewidth]{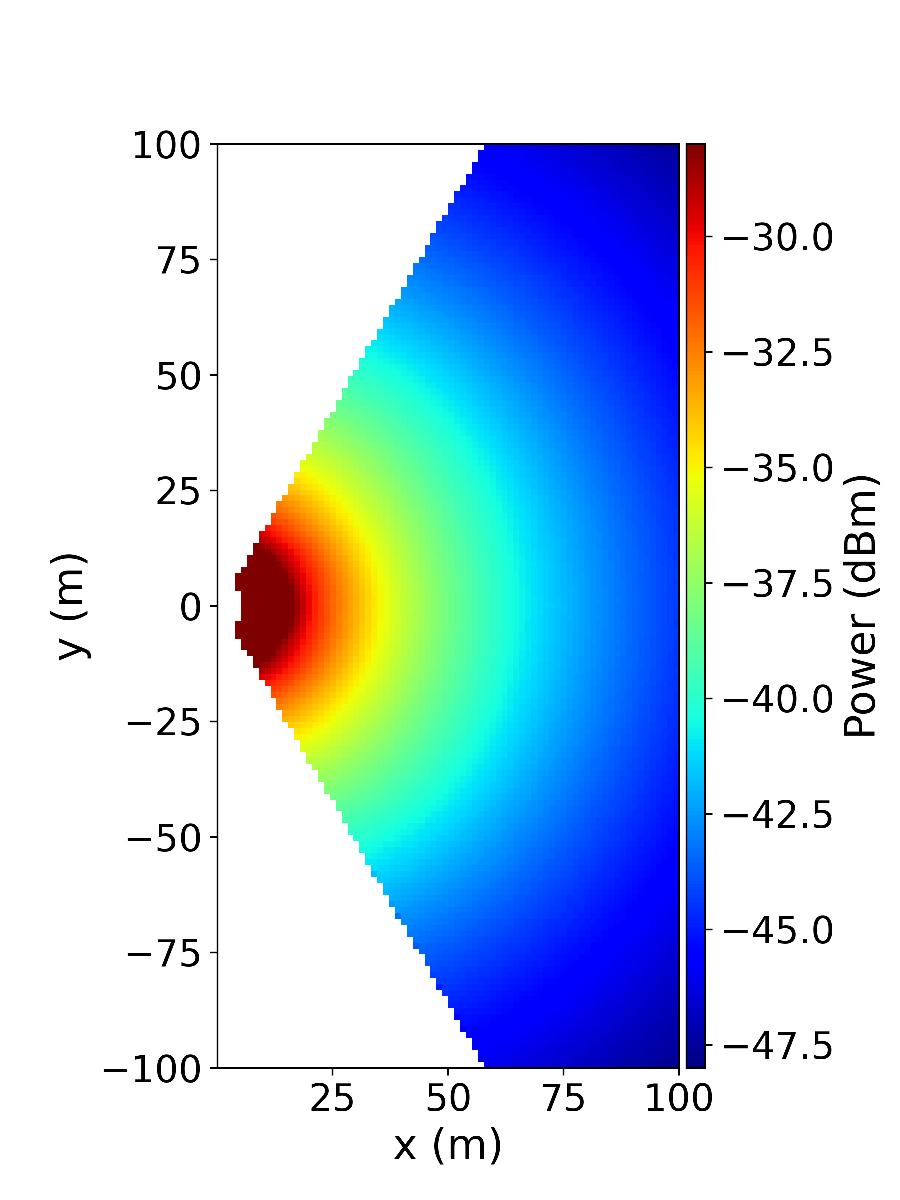}
        \label{fig: spot_5_50}
    }
    \hfil
    \subfloat[SPOT beam (5--300 m)]{
        \includegraphics[width=0.39\linewidth]{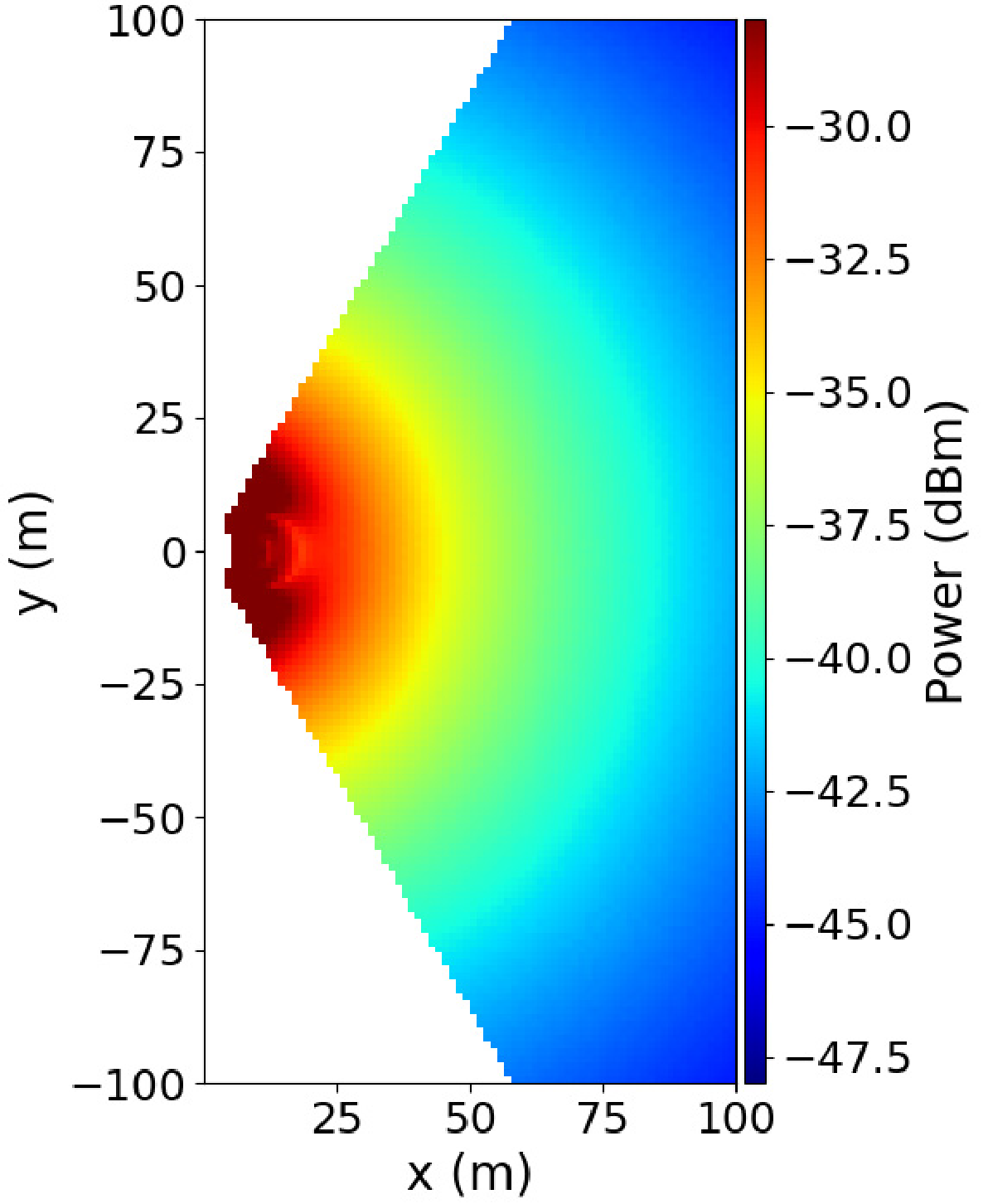}
        \label{fig: spot_5_300}
    }
    \caption{Maximum received power of rainbow beams (dBm).}
    \label{fig: beampattern}
    \vspace{-1.8em}
\end{figure}

Next, we show how the SPOT beam is a better choice for distance and position estimation. Fig.~\ref{fig: beampattern} illustrates the maximum received power of rainbow beams generated by the CBS baseline and the proposed SPOT design under two user-range priors: a near-range distribution (5–50 m) and a wide-range distribution (5–300 m). The CBS beams for angle estimation (CBS Beam 1) in Fig.~\ref{fig: cbs_5_50} and Fig.~\ref{fig: cbs_5_300} are configured to span from $(-60^{\circ},10~\text{m})$ to $(60^{\circ},10~\text{m})$ and from $(-60^{\circ},200~\text{m})$ to $(60^{\circ},200~\text{m})$, respectively. 
The SPOT beams in Fig.~\ref{fig: spot_5_50} and Fig.~\ref{fig: spot_5_300} maintain an almost monotonic power–distance relationship over all angles, offering a stable foundation for power-based ranging. Therefore, the learned beam acts as a spatial encoder: the subcarrier index encodes the angle and the received power encodes the distance, allowing both parameters to be inferred from a single beam.
Moreover, SPOT dynamically adapts its power–distance profile to the user range: in near-range scenarios, pronounced power variations enable fine distance discrimination, whereas in wide-range scenarios, a deliberately smoother profile mitigates error accumulation over extended distances, attaining better ranging accuracy than the CBS baseline.

\subsection{2D RMSE under Different Quantization Bit Lengths}
\begin{figure}[t]    
    \centering    
    \includegraphics[width=1\linewidth]{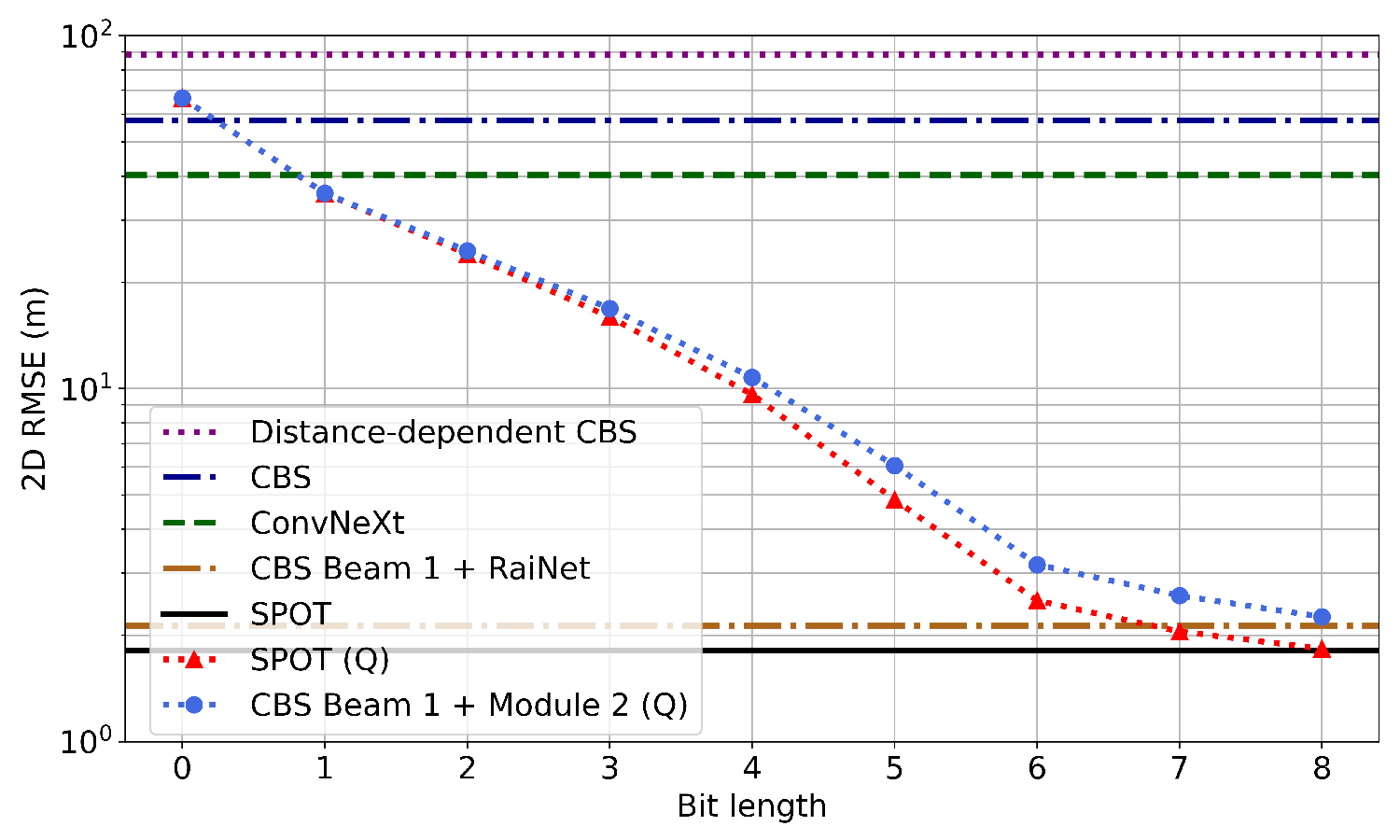}
    \caption{2D RMSE under different power quantization bit length.}
    \label{fig: bit_length}
\vspace{-1.5 em}
\end{figure}
Fig.~\ref{fig: bit_length} depicts the effect of power-feedback quantization bit length for users distributed in $[5,300]$~m. The 2D RMSE of the quantized case `SPOT(Q)' decreases exponentially as the bit length increases from 1 to 5, and the performance at 8 bits nearly matches that of unquantized feedback. We also evaluate a scheme with a fixed CBS beam and trained estimator (`CBS beam 1 + Module 2 (Q)'). Results show that the joint training approach SPOT(Q) achieves lower RMSE, with the performance gap widening as bit length increases. Specifically, at 8-bit quantization versus the fixed-beam strategy, joint training cuts RMSE by 18.67\% from 2.26 m to 1.83 m.



\subsection{Computational Complexity and Feedback Overhead}
\begin{small}
\begin{table*}[htbp]
    \centering
    \caption{Complexity and overhead comparison. FP32 precision is assumed for all floating-point signal representations. }
    \label{tab:comparison}
    \begin{tabular}{|c|c|c|c|c|c|}
        \hline
       \textbf{Algorithm} & \textbf{FLOPs} & \makecell{\textbf{Downlink}\\ \textbf{Beams}} & \makecell{\textbf{Uplink Feedback} \\ \textbf{Rounds per User}} & \makecell{\textbf{Total Feedback}\\ \textbf{Overhead per User (bits)}} & \textbf{Feedback Contents per User} \\
        \hline
        CBS \cite{luo2024beam} & - & $1+K$ & $2$ &  $22$ & two 11-bit($\lceil \log_2 1584\rceil$) indices \\
        \hline
        ConvNeXt \cite{lei2025deep} & 4.98M  & $1+K$ & $2$ &  $\num{202752}$ & two length-$1584$ full signal vectors \\
        \hline
        Distance-dependent CBS \cite{zheng2025near} & -  & $10$ & $1$ &  $14$ & 14-bit($\lceil \log_2 15840\rceil$) index \\
        \hline
        RaiNet \cite{Klus2025Deep} & 19.37M  & $1$ & $1$ &  $\num{50688}$ & length-$1584$ signal power vector \\
        \hline
        SPOT & 0.035M  & 1 & $1$ & \(
        19\) & $11$-bit index and $8$-bit quantized power value \\
        \hline
    \end{tabular}
\vspace{-1.em}
\end{table*}
\end{small}

Table~\ref{tab:comparison} compares the complexity and overhead of SPOT and baselines. Both CBS and distance-dependent CBS incur minimal complexity, as they directly infer user location from precomputed beam-splitting trajectories. In contrast, ConvNeXt refines a coarse estimate using a convolutional network that processes the full received-signal vector, greatly increasing computational and feedback costs. Moreover, CBS and ConvNeXt require two downlink transmissions with separate feedback for angle and distance estimation. For distance-dependent CBS, the \ac{BS} sequentially transmits 10 distance-aware rainbow beams to sweep the entire sector, and the user reports back the subcarrier index corresponding to the strongest received power observed across all rounds. RaiNet requires only a single downlink transmission but incurs a high uplink cost due to the feedback of the full power vector and a larger network complexity of 19.37 M FLOPs. In contrast, SPOT attains the best trade-off between accuracy and efficiency, requiring merely 0.035 M FLOPs with one downlink transmission and a single uplink message containing only two scalar values.

\section{Conclusion}
In this paper, we proposed a novel deep learning scheme for wideband \ac{ISAC} systems employing \acp{PTA}. The core innovation lies in modeling \ac{PS} and \ac{TTD} coefficients as learnable parameters within the network, enabling joint optimization of rainbow beam generation and position estimation. Using only a single downlink transmission and compact uplink feedback containing the quantized maximum received power and its subcarrier index, the proposed method efficiently recovers both angle and distance, greatly reducing signaling overhead. Extensive simulations across a wide range of user distances demonstrate that our scheme attains lower localization RMSE with minimal cost. The performance advantage becomes more pronounced at larger user distances, demonstrating strong robustness across diverse propagation conditions. 


\bibliographystyle{IEEEtran}
\bibliography{reference}

\end{document}

%% file: acronyms.tex
\DeclareAcronym{3GPP}{
  short=3GPP,
  long=3rd generation partnership project
}
\DeclareAcronym{ADC}{
  short=ADC,
  long=analog-to-digital converter
}
\DeclareAcronym{AMP}{
  short=AMP,
  long=approximate message passing
}
\DeclareAcronym{ANN}{
  short=ANN,
  long=artificial neural network
}
\DeclareAcronym{AoA}{
  short=AoA,
  long=angle-of-arrival
}
\DeclareAcronym{AoD}{
  short=AoD,
  long=angle-of-departure
}
\DeclareAcronym{APS}{
  short=APS,
  long=azimuth power spectrum
}
\DeclareAcronym{PS}{
  short=PS,
  long=phase shifter,
  long-plural-form = phase shifters
}
\DeclareAcronym{AR}{
  short=AR,
  long=augmented reality
}
\DeclareAcronym{AV}{
  short=AV,
  long=autonomous vehicle
}
\DeclareAcronym{BM}{
  short=BM,
  long=beam management
}
\DeclareAcronym{BS}{
  short=BS,
  long=base station
}
\DeclareAcronym{BSM}{
  short=BSM,
  long=basic safety message
}
\DeclareAcronym{BW}{
  short=BW,
  long=bandwidth
}
\DeclareAcronym{CDF}{
  short=CDF,
  long=cumulative distribution function
}
\DeclareAcronym{CP}{
  short=CP,
  long=cyclic-prefix
}
\DeclareAcronym{CSI-RS}{
  short=CSI-RS,
  long=channel state information reference signal
}
\DeclareAcronym{DFT}{
  short=DFT,
  long=discrete Fourier transform
}
\DeclareAcronym{DL}{
  short=DL,
  long=downlink
}
\DeclareAcronym{EKF}{
  short=EKF,
  long=extended Kalman filter
}
\DeclareAcronym{DSRC}{
  short=DSRC,
  long=dedicated short-range communication
}
\DeclareAcronym{FDD}{
  short=FDD,
  long=frequency division duplex
}
\DeclareAcronym{FMCW}{
  short=FMCW,
  long=frequency modulated continuous wave
}
\DeclareAcronym{FoV}{
  short=FoV,
  long=field-of-view
}
\DeclareAcronym{GNSS}{
  short=GNSS,
  long=global navigation satellite system
}

\DeclareAcronym{ISAC}{
    short=ISAC,
    long=Integrated sensing and communication
}

\DeclareAcronym{IMU}{
  short=IMU,
  long=inertial measurement unit
}
\DeclareAcronym{lidar}{
  short=lidar,
  long=light detection and ranging
}
\DeclareAcronym{LOS}{
  short=LOS,
  long=line-of-sight
}
\DeclareAcronym{LPF}{
  short=LPF,
  long=low pass filter
}
\DeclareAcronym{LTE}{
  short=LTE,
  long=long term evolution
}
\DeclareAcronym{MIMO}{
  short=MIMO,
  long=multiple-input multiple-output
}
\DeclareAcronym{ML}{
  short=ML,
  long=machine learning
}
\DeclareAcronym{mmWave}{
  short=mmWave,
  long=millimeter wave
}
\DeclareAcronym{MRR}{
  short=MRR,
  long=medium range radar
}
\DeclareAcronym{NLOS}{
  short=NLOS,
  long=non-line-of-sight
}
\DeclareAcronym{NB}{
  short=NB,
  long=narrow beam
}
\DeclareAcronym{NR}{
  short=NR,
  long=new radio
}
\DeclareAcronym{OFDM}{
  short=OFDM,
  long=orthogonal frequency-division multiplexing
}
\DeclareAcronym{ppm}{
  short=ppm,
  long=parts-per-million
}
\DeclareAcronym{PF}{
  short=PF,
  long=particle filter
}
\DeclareAcronym{RMS}{
  short=RMS,
  long=root-mean-square
}
\DeclareAcronym{RPE}{
  short=RPE,
  long=relative precoding efficiency
}
\DeclareAcronym{RS}{
  short=RS,
  long=reference signal
}
\DeclareAcronym{RSRP}{
  short=RSRP,
  long=reference signal received power
}
\DeclareAcronym{RSU}{
  short=RSU,
  long=roadside unit
}
\DeclareAcronym{SCS}{
  short=SCS,
  long=subcarrier spacing
}
\DeclareAcronym{SNR}{
  short=SNR,
  long=signal-to-noise ratio
}
\DeclareAcronym{SSB}{
  short=SSB,
  long=synchronization signal block
}
\DeclareAcronym{THz}{
  short=THz,
  long=terahertz
}
\DeclareAcronym{TTD}{
  short=TTD,
  long=true-time delay,
  long-plural-form = true-time delays
}

\DeclareAcronym{UAV}{
  short=UAV,
  long=unmanned aerial vehicle
}
\DeclareAcronym{UE}{
  short=UE,
  long=user equipment
}
\DeclareAcronym{UKF}{
  short=UKF,
  long=unscented Kalman filter
}
\DeclareAcronym{RF}{
  short=RF,
  long=radio frequency
}
\DeclareAcronym{UL}{
  short=UL,
  long=uplink
}
\DeclareAcronym{ULA}{
  short=ULA,
  long=uniform linear array
}
\DeclareAcronym{V2I}{
  short=V2I,
  long=vehicle-to-infrastructure
}
\DeclareAcronym{V2V}{
  short=V2V,
  long=vehicle-to-vehicle
}
\DeclareAcronym{V2X}{
  short=V2X,
  long=vehicle-to-everything
}
\DeclareAcronym{VR}{
  short=VR,
  long=virtual reality
}
\DeclareAcronym{VRU}{
  short=VRU,
  long=vulnerable road user
}
\DeclareAcronym{WB}{
  short=WB,
  long=wide beam
}
\DeclareAcronym{RNN}{
	short=RNN,
	long=recurrent neural network
}
\DeclareAcronym{LSTM}{
	short=LSTM,
	long=long short-term memory
}
\DeclareAcronym{LoS}{
	short=LoS,
	long=line-of-sight
}
\DeclareAcronym{FC}{
	short=FC,
	long=fully connected
}

\DeclareAcronym{PC}{
	short=FC,
	long=partially connected
}
\DeclareAcronym{PTA}{
  short = PTA,
  long  = phase-time array,
  long-plural-form = phase-time arrays
}

\DeclareAcronym{PAA}{
    short=PAA,
    long = phased antenna arrays
}